\documentclass{article}

\usepackage{arxiv}

\usepackage[utf8]{inputenc} % allow utf-8 input
\usepackage[colorlinks=true,breaklinks=true,linkcolor=blue,citecolor=blue,urlcolor=blue]{hyperref} % hyperlinks
\usepackage{url}            % simple URL typesetting
\usepackage{booktabs}       % professional-quality tables
\usepackage{amsfonts}       % blackboard math symbols
\usepackage{amsmath,amssymb}
\usepackage{graphicx}
\usepackage{natbib}
\usepackage{float}
\usepackage{xcolor}
\usepackage{mdframed}
\usepackage{algorithm}
\usepackage{algpseudocode}

\usepackage{amsmath,amsfonts,bm}

\def\eqref#1{equation~\ref{#1}}
\def\1{\bm{1}}

\DeclareMathAlphabet{\mathsfit}{\encodingdefault}{\sfdefault}{m}{sl}
\SetMathAlphabet{\mathsfit}{bold}{\encodingdefault}{\sfdefault}{bx}{n}

\newcommand{\op}[1]{\textsc{#1}}

\title{RewardWeaver: Long-Horizon Interactive Learning for Language Agents\\
via Self-Evolving Reward Adaptation}

\author{%
  Hengbo Xiao\textsuperscript{$\ast\spadesuit$} \\
  Yoolee.ai \\
  \And
  Boyao Zhang\textsuperscript{$\ast$} \\
  University of Science and Technology of China \\
  \And
  Purui Liu\textsuperscript{$\ast$} \\
  Peking University \\
  \And
  Yuxuan Zheng \\
  Peking University \\
  \And
  Haoran Yin \\
  Tianjin University \\
  \And
  Haibo Liu \\
  Yoolee.ai \\
  \And
  Fan Zhang\textsuperscript{$\dagger$} \\
  Yoolee.ai \\
}

\date{\texttt{fan.zhang@yoolee.ai}}

\renewcommand{\shorttitle}{RewardWeaver: Self-Evolving Reward Adaptation}

\hypersetup{
pdftitle={RewardWeaver: Long-Horizon Interactive Learning for Language Agents via Self-Evolving Reward Adaptation},
pdfsubject={cs.CL, cs.AI, cs.LG},
pdfauthor={Hengbo Xiao, Boyao Zhang, Purui Liu, Yuxuan Zheng, Haoran Yin, Haibo Liu, Fan Zhang},
pdfkeywords={reward adaptation, reinforcement learning from verifiable rewards, language agents, long-horizon interaction, process rewards},
}

\begin{document}

\maketitle
% Equal-contribution / corresponding-author note, emitted as a footnote
% without a visible marker so no extra symbol prints after Fan Zhang's
% dagger.
{\let\thefootnote\relax\footnotetext{$\ast$ Equal contribution.
$\spadesuit$ Project leader. $\dagger$ Corresponding author.}}

% Abstract and Introduction for the RewardWeaver manuscript.

\begin{abstract}
Reinforcement learning with verifiable rewards (RLVR) has driven
substantial progress in domains where task outcomes can be reliably
evaluated, but long-horizon interaction remains challenging due to sparse
terminal feedback and difficult credit assignment. Process rewards provide
denser supervision, yet the capabilities most relevant for training can
change as the policy evolves: a behavior that is easy to evaluate or
frequently deficient need not be the bottleneck currently limiting task
success. We introduce \textbf{RewardWeaver}, a self-evolving reward
adaptation framework for language agents in long-horizon interaction.
RewardWeaver maintains a validated capability space in which the
semantics of admitted Rubrics remain fixed, and closes the loop between
policy optimization, task evaluation, failure attribution, and reward
adaptation. After each training stage, it performs
outcome-grounded backward attribution on low-outcome trajectories,
aggregates recurrent and policy-controlled capability bottlenecks, and
dynamically selects the corresponding process rewards for the
next stage. Recurrent failures not covered by the existing capability space
trigger a separate, controlled expansion procedure. We evaluate
\textsc{RewardWeaver} on SOTOPIA, AmazonHistoryPrice, and a newly
constructed Sales Benchmark. Across social
interaction, bilateral bargaining, and domain-specific sales,
\textsc{RewardWeaver} establishes new state-of-the-art (SOTA)
results. Ablations further demonstrate the importance
of dynamic reward allocation, failure-grounded attribution, and stable
semantics for admitted capabilities.
\end{abstract}

\section{Introduction}
\label{sec:introduction}

Reinforcement learning with verifiable rewards (RLVR) has enabled
substantial progress in mathematics and code, where task
outcomes can be reliably checked by rules, tests, or execution
\citep{rlvr,rlvr_code,deepseek_r1}. Long-horizon interaction,
however, poses a fundamentally different optimization problem. In settings such as
negotiation, social coordination, and collaborative
decision making \citep{sotopia,ampo,epo}, each action changes the partner's
subsequent behavior and
reshapes the interaction trajectory. Final outcomes emerge from
sequences of interdependent decisions across many turns, making rewards
sparse, delayed, and difficult to assign to specific behaviors
\citep{rudder,tvt,cocoa}. A terminal
score indicates whether an interaction succeeded, but certifies failure
without localizing its cause and cannot by itself direct training.

\begin{figure}[t]
    \centering
    \includegraphics[width=0.9\textwidth]{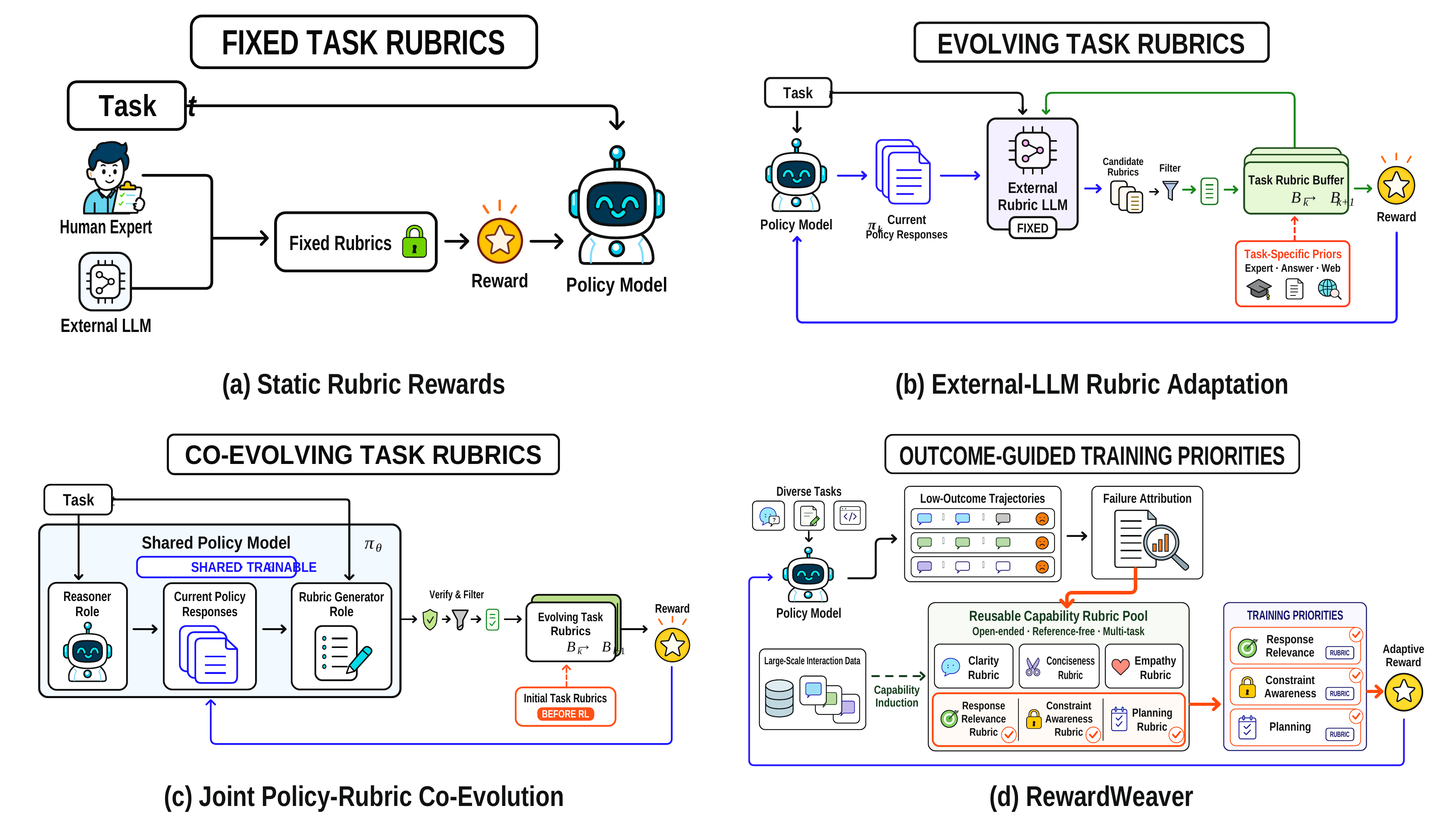}
    \caption{\textbf{Reward Adaptation under Policy Evolution.} \textbf{(a)} Task-specific Rubrics remain fixed throughout training. \textbf{(b)} External-LLM methods revise Rubrics as policy evolves. \textbf{(c)} Co-evolution methods jointly update Rubrics and the policy. \textbf{(d)} RewardWeaver grounds reward adaptation in recurrent task failures and prioritizes capability Rubrics according to current policy bottlenecks.}
    \label{fig:reward-adaptation-paradigms}
\end{figure}

Rubric-based rewards and evaluators provide denser supervision by
assessing intermediate behaviors against explicit, fine-grained criteria
\citep{rar,qurl,llm_rubric,checkeval}, and rubric construction now scales
from instance-specific generation to reusable repositories
\citep{advancedif,rubrichub}. Recent approaches further
revise or evolve these criteria as the policy changes, tracking its
current response distribution
\citep{evorubric}. However, tracking the response distribution is not
tracking the failure distribution: a behavior may be frequent or
easy to discriminate without being a recurrent cause of task failure,
while a less frequent capability deficiency may constitute the actual
performance bottleneck. The key question is not only how
evaluation criteria should adapt, but what evidence should drive that
adaptation. We argue that this evidence should be recurrent failures
in low-outcome trajectories, rather than response
characteristics alone.
Figure~\ref{fig:reward-adaptation-paradigms} contrasts these
reward-adaptation paradigms.

The central challenge is not merely how to design better process
rewards, but how to decide what the current policy should optimize next.
This introduces an outer-loop optimization problem complementary to
standard RL: the inner-loop optimizer determines how to improve the policy
under a given reward, while the outer loop must determine which
capabilities deserve training credit. Diagnostic evidence must
therefore satisfy three conditions: it must be outcome-grounded, drawn
from trajectories that failed rather than sampled uniformly; recurrent,
reappearing across trajectories rather than reflecting incidental
variation; and policy-controlled, attributable to the agent's own
behavior \citep{failure_attribution} rather than to the partner or the
scenario. Isolated low scores satisfy none of these conditions. In this
view, the task-level
objective should remain stable, while process-level training credit shifts
toward the capabilities that currently constrain performance.

We introduce \textbf{RewardWeaver}, a self-evolving reward adaptation
framework for language agents in long-horizon interaction.
\textsc{RewardWeaver} closes the loop between
policy training, automatic evaluation, failure attribution, and reward
adaptation. At each stage, it identifies recurrent capability bottlenecks
from on-policy interaction failures and dynamically selects the
corresponding Rubrics to construct the next process reward. Unlike prior self-evolving rubric methods that evolve \emph{what is
evaluated}, \textsc{RewardWeaver} evolves \emph{what is optimized}: it
preserves stable semantics for admitted capabilities while dynamically
reallocating training credit across them. Recurrent failures that fall outside the existing Rubric
space are incorporated only through controlled validation, allowing the
reward to evolve without uncontrolled semantic drift.

We evaluate \textsc{RewardWeaver} on SOTOPIA \citep{sotopia} and related
social intelligence benchmarks against outcome-only reinforcement learning,
static capability prioritization, and alternative attribution and
rubric-design strategies.
\textsc{RewardWeaver}
improves the primary evaluation metrics and achieves new
state-of-the-art (SOTA) performance under matched training and evaluation
protocols. Further ablation, sensitivity, and
qualitative analyses show that failure-grounded attribution, adaptive
credit reallocation, and stable semantics for admitted capabilities are
key contributors to the gains. Our
contributions are summarized as follows: (1) We formulate reward adaptation
in long-horizon interaction as an outer-loop
optimization problem that determines which process capabilities should be
prioritized next. (2) We propose \textsc{RewardWeaver}, which dynamically
reallocates training credit through failure-driven capability-bottleneck
identification while preserving stable task objectives and evaluation
semantics. (3) We conduct systematic experiments demonstrating the
advantages of self-evolving reward adaptation over outcome-only and static
process-reward baselines, and isolate the contributions of
\textsc{RewardWeaver}'s key components.

\section{RewardWeaver Methodology}
\label{sec:method}

\begin{figure*}[t]
    \centering
    \includegraphics[width=\textwidth]{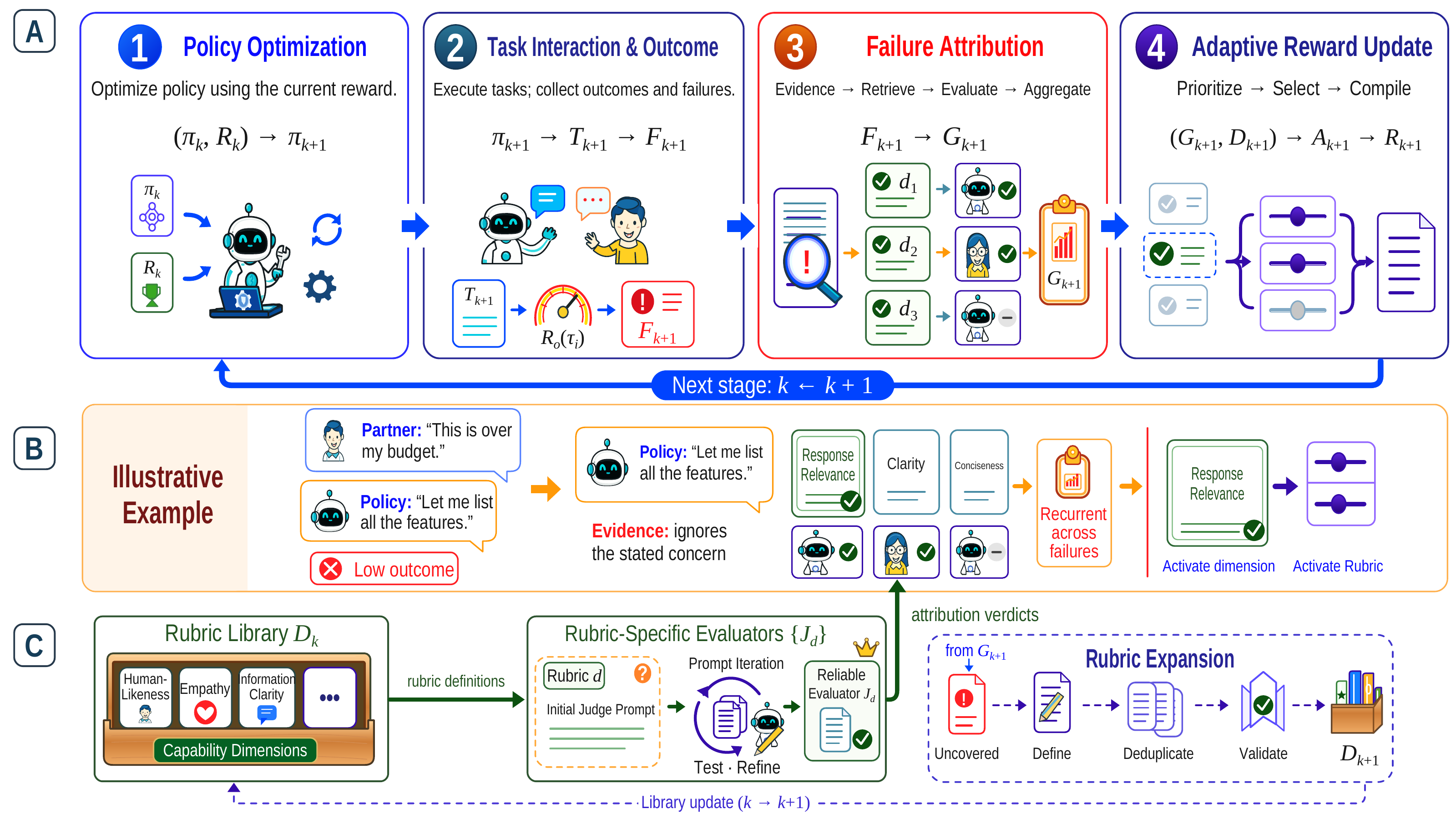}
    \caption{
    \textbf{The Overall Architecture of \textsc{RewardWeaver}.}
    \textbf{(a)} The policy is optimized under the current reward; its
    on-policy low-outcome trajectories reveal recurrent capability
    bottlenecks that determine the next-stage allocation of training
    credit.
    \textbf{(b)} Failure attribution grounds bottleneck identification in
    concrete behavioral evidence.
    \textbf{(c)} A validated capability space provides stable semantics for
    admitted capabilities, while recurrent uncovered failures trigger
    controlled expansion.
    }
    \label{fig:autorl-overview}
\end{figure*}

\textsc{RewardWeaver} separates \emph{reward semantics} from
\emph{reward relevance}: a validated capability space determines
\emph{what can be rewarded}, while a stage-wise outer loop determines
\emph{what should be rewarded}. Training priorities can evolve
with the policy without the semantics of admitted capabilities drifting,
while the capability library itself expands only through a separate
validated procedure.

% ============================================================
\subsection{Task Formulation}
\label{sec:problem-setup}

At stage $k$, policy $\pi_k$ interacts with partner
$\pi_{\mathrm{partner}}$: given interaction history $h_t$, the two parties
act alternately with $a_t\sim\pi_k(\cdot\mid h_t)$ and
$b_t\sim\pi_{\mathrm{partner}}(\cdot\mid h_t,a_t)$.
A completed trajectory $\tau$ receives a fixed task-level outcome
$R_o(\tau)$ and a stage-dependent process reward,
\begin{equation}
R_k(\tau)
=
R_o(\tau)
+
\frac{1}{|A_k|}
\sum_{d\in A_k}
\bar J_d(\tau),
\label{eq:stage-reward}
\end{equation}
where $\mathcal D_k$ is the capability space available at stage $k$,
$A_k\subseteq\mathcal D_k$ the active subset, and $\bar J_d(\tau)$ the
trajectory-level score of capability evaluator $J_d$; all active Rubrics
enter with equal weight, so adaptation determines \emph{which}
capabilities receive credit rather than how much each receives. The terminal
objective remains fixed; only process-level credit changes: standard RL
determines \emph{how} to optimize under $R_k$, i.e.,
$(\pi_k,R_k)\rightarrow\pi_{k+1}$,
whereas \textsc{RewardWeaver} determines
\emph{which capabilities should receive training credit next}.
Throughout, \emph{capability} denotes the behavioral concept,
\emph{Rubric} its frozen operational definition, and \emph{evaluator} the
scoring function $J_d$; the \emph{active set} is the subset of
capabilities currently receiving process-level training credit, and a
\emph{reward specification} couples the active set with how its
evaluators are composed.

% ============================================================
\subsection{Validated Capability Space}
\label{sec:rubric-library}

Reward adaptation requires stable semantics for admitted
capabilities. \textsc{RewardWeaver} operates over a validated space
$\mathcal D$ of observable, policy-controllable behaviors, where each
$d\in\mathcal D$ couples a frozen Rubric definition with an
evaluator $J_d(x)=M_{\mathrm{eval}}(p_d,x)$, and $p_d$ operationalizes
the definition over interaction context $x$. Construction has two stages: candidates are
first induced from interaction evidence, consolidated across runs, and
filtered for generality, operationality, and semantic distinctiveness;
each is then operationalized by a dedicated evaluator and subjected to
source-disjoint acceptance testing (Algorithm~\ref{alg:capability-space}).
Candidates may be revised before admission; only validated pairs are
admitted into $\mathcal D$, so discovery remains flexible while admitted
semantics remain fixed. Admission testing certifies each evaluator on its
own source-disjoint acceptance set, so every admitted capability satisfies
the same validity criteria. The SOTOPIA capability space, representative
entries, and their operationalization into executable criteria are
provided in Appendix~\ref{app:rubric-library}.

\begin{algorithm}[t]
\caption{\textbf{Validated Capability-Space Construction}}
\label{alg:capability-space}
\small
\begin{algorithmic}[1]

\Require Raw interaction corpus $\mathcal T_{\mathrm{raw}}$,
induction models $\mathcal M$,
validity criteria $\mathcal V$,
evaluator configuration $\kappa$,
repeat count $R$,
construction budget $B$

\Ensure Validated capability space
$\mathcal D=\{(d,J_d)\}$

\Statex \textbf{Stage I: Capability Discovery and Consolidation}

\State $\mathcal T
\gets
\op{NormalizeAndFilter}(\mathcal T_{\mathrm{raw}})$

\State $\mathcal F
\gets
\op{OpenInductionWithEvidence}(\mathcal T,\mathcal M)$

\For{$r=1,\ldots,R$}
    \State $\mathcal S_r
    \gets
    \op{Consolidate}
    (\op{EmbeddingNeighbors}(\mathcal F))$

    \State $\mathcal S_r
    \gets
    \op{Filter}
    (\mathcal S_r;
    \text{generality},
    \text{operationality},
    \text{distinctiveness})$
\EndFor

\State $\mathcal S
\gets
\op{AlignAndSelect}
(\{\mathcal S_r\}_{r=1}^{R};
\text{recurrence},
\text{coverage},
\text{support})$

\State $\mathcal S
\gets
\op{DefineCandidates}(\mathcal S,\mathcal F)$

\Statex \textbf{Stage II: Operationalization and Admission}

\State $\mathcal D\gets\varnothing$

\While{$\mathcal S\neq\varnothing$ and budget $B$ remains}

    \State remove candidate $d$ from $\mathcal S$

    \State $\mathcal X_d
    \gets
    \op{RetrieveRelevantCases}(\mathcal T,d)$

    \State $\mathcal L_d
    \gets
    \op{AnnotateAndVerify}(\mathcal X_d,d)$

    \State $(\mathcal L_d^{\mathrm{dev}},
            \mathcal L_d^{\mathrm{acc}})
    \gets
    \op{SourceDisjointSplit}(\mathcal L_d)$

    \State $(\mathcal L_d^{\mathrm{dev}},
            \mathcal L_d^{\mathrm{acc}})
    \gets
    \op{EnsureCoverage}
    (\mathcal L_d^{\mathrm{dev}},
     \mathcal L_d^{\mathrm{acc}},d)$

    \State $p_d^\star
    \gets
    \op{DevelopEvaluator}
    (d,\mathcal L_d^{\mathrm{dev}},\kappa)$

    \State $J_d
    \gets
    \op{FreezeEvaluator}(p_d^\star,d,\kappa)$

    \State $v_d
    \gets
    \op{ValidateCandidate}
    (d,J_d,\mathcal L_d^{\mathrm{acc}},
     \mathcal D;\mathcal V)$

    \If{$v_d=\textsc{Accept}$}

        \State $\mathcal D
        \gets
        \mathcal D
        \cup
        \{(\op{FreezeDefinition}(d),J_d)\}$

    \ElsIf{$v_d=\textsc{Revise}$}

        \State $d'
        \gets
        \op{ReviseCandidate}(d,v_d)$

        \State $\mathcal S
        \gets
        \mathcal S\cup\{d'\}$

    \Else

        \State discard $d$

    \EndIf
\EndWhile

\State \Return $\mathcal D$

\end{algorithmic}
\end{algorithm}

% ============================================================
\subsection{Outcome-Grounded Failure Attribution}
\label{sec:failure-attribution}

Terminal outcomes identify low-outcome trajectories that warrant
diagnosis, but do not by themselves establish an agent-side failure or
determine what should receive training credit. After stage $k$, $\pi_{k+1}$ generates
on-policy trajectories $\mathcal T_{k+1}$, from which the low-outcome
cohort
$\mathcal F_{k+1}
=
\{\tau_i\in\mathcal T_{k+1}
\mid
q_o(R_o(\tau_i),x_i)=1\}$
is selected, where $x_i$ denotes the task instance of
trajectory $\tau_i$. For each $\tau_i\in\mathcal F_{k+1}$,
\textsc{RewardWeaver} localizes failure-relevant evidence, retrieves
candidate capabilities $C_i\subseteq\mathcal D_k$, and produces
attribution records $a_{i,d}=(s_{i,d},c_{i,d},e_{i,d})$ for $d\in C_i$,
where $s_{i,d}$ is the deficit judgment, $c_{i,d}$ its confidence, and
$e_{i,d}$ the supporting evidence; unsupported assignments are rejected.
Trajectory-level records are aggregated into
$G_{k+1}=\operatorname{Aggregate}(\{a_{i,d}\})$,
which retains recurrent, supported, and policy-controlled
failure modes rather than isolated low scores.

% ============================================================
\subsection{Adaptive Credit Allocation}
\label{sec:reward-adaptation}

A capability can be valid yet irrelevant to the current policy;
\textsc{RewardWeaver} treats training relevance as stage-dependent.
Given $G_{k+1}$, it selects a budgeted active set
$A_{k+1}=\operatorname{Select}(G_{k+1},\mathcal D_k)
\subseteq\mathcal D_k$.
All active Rubrics contribute equally to the process reward,
yielding
$R_{k+1}(\tau)
=
R_o(\tau)
+
\frac{1}{|A_{k+1}|}
\sum_{d\in A_{k+1}}
\bar J_d(\tau)$.
As a bottleneck recedes, its Rubric may leave the active set and
capacity is released for newly dominant failures, so the process
reward is selective and stage-dependent rather than cumulative.
\begin{algorithm}[t]
\caption{\textbf{Failure Attribution and Adaptive Credit Allocation}}
\label{alg:attribution-adaptation}
\small
\begin{algorithmic}[1]

\Require On-policy trajectories $\mathcal T_{k+1}$,
capability space $\mathcal D_k$,
current reward specification $P_k$,
adaptation history $\mathcal S_k$,
task-performance summary $y_{k+1}$,
calibration set $\mathcal H_{k+1}$,
parameters $\Theta$ with active budget $K$
and calibration limit $M$

\Ensure Next-stage reward specification $P_{k+1}$,
or calibration failure $\bot$

\State $\mathcal F_{k+1}\gets\op{LowOutcomes}(\mathcal T_{k+1})$;
\quad $\mathcal Z_{k+1}\gets\varnothing$

\For{$\tau\in\mathcal F_{k+1}$}
    \State $e_\tau\gets\op{LocalizeEvidence}(\tau)$
    \State $C_\tau\gets\op{Retrieve}(e_\tau,\mathcal D_k)$
    \State $\mathcal Z_\tau\gets\op{Attribute}(e_\tau,C_\tau)$
    \Comment{Abstention allowed}
    \State $\mathcal Z_{k+1}
    \gets
    \mathcal Z_{k+1}\cup\mathcal Z_\tau$
\EndFor

\State $(G_{k+1},\{r_d,n_d,c_d,f_d\})
\gets
\op{Aggregate}(\mathcal Z_{k+1})$

\State $p_d
\gets
\alpha r_d+\beta c_d+\gamma(1-f_d)$,
\quad $d\in\mathcal D_k$

\State $(\mathcal B^{\mathrm{high}},
        \mathcal B^{\mathrm{weak}})
\gets
\op{Eligible}(G_{k+1},\mathcal S_k;\Theta)$

\State $A_{k+1}
\gets
\op{TopK}(\mathcal B^{\mathrm{high}},p,K)$

\If{$|A_{k+1}|<K$
    \textbf{and}
    $\op{AllowWeak}(\mathcal B^{\mathrm{high}},
    \mathcal S_k;\Theta)$}

    \State $A_{k+1}
    \gets
    A_{k+1}
    \cup
    \op{TopK}
    (\mathcal B^{\mathrm{weak}}\setminus A_{k+1},
    p,K-|A_{k+1}|)$
\EndIf

\State $(A_{k+1},\mathcal S_{k+1})
\gets
\op{Schedule}
(A_{k+1},G_{k+1},\mathcal S_k,y_{k+1};\Theta)$

\State $\lambda_{k+1,d}
\gets
1/|A_{k+1}|$,
\quad $d\in A_{k+1}$
\Comment{Equal weight over active Rubrics}

\State $Q_{k+1}
\gets
\op{CompileRubrics}(A_{k+1},P_k)$
\Comment{Assemble a reward specification}

\For{$m=1,\ldots,M$}

    \State $(s,\delta)
    \gets
    \op{Calibrate}(Q_{k+1},\mathcal H_{k+1})$

    \If{$\op{Accept}(s)$}
        \Comment{Rubric-set publication gate}

        \State $P_{k+1}
        \gets
        \op{Compose}
        (P_k,A_{k+1},\lambda_{k+1},
        Q_{k+1},s,\delta,\mathcal S_{k+1})$

        \State $\op{Publish}(P_{k+1})$
        \State \Return $P_{k+1}$

    \ElsIf{$m<M$}

        \State $Q_{k+1}
        \gets
        \op{Revise}(Q_{k+1},s,\delta)$
        \Comment{Revise composition; keep $(d,J_d)$ frozen}

    \EndIf

\EndFor

\State \Return $\bot$
\Comment{Calibration failure; no new reward is published}

\end{algorithmic}
\end{algorithm}
Letting $P_k$ denote the deployed process-reward specification at
stage $k$, $\mathcal S_k$ its adaptation history, and $y_{k+1}$ the
stage-level task-performance summary,
Algorithm~\ref{alg:attribution-adaptation} specifies the complete
attribution-to-reward update.
Importantly, the update changes which Rubrics are active and how
they are composed into the next-stage reward, but never modifies
the frozen semantics or evaluators of the capabilities admitted to
$\mathcal D_k$.

% ============================================================
\subsection{Controlled Capability Expansion}
\label{sec:capability-expansion}

Adaptive credit allocation operates only over the capability space already
validated at the beginning of the update, so the transition from stage $k$
to $k+1$ always satisfies
$A_{k+1}\subseteq\mathcal D_k$.
When recurrent failures cannot be represented by $\mathcal D_k$,
\textsc{RewardWeaver} invokes the same admission procedure as
Algorithm~\ref{alg:capability-space}: validated additions form the
expanded space $\mathcal D_{k+1}$, but newly admitted capabilities do not
alter the reward update that triggered their discovery—they become
eligible only in the subsequent adaptation step, i.e.,
$A_{k+2}\subseteq\mathcal D_{k+1}$.
The fast loop therefore reallocates training credit within fixed
semantics,
\begin{equation}
(\pi_k,R_k)
\rightarrow
\pi_{k+1}
\rightarrow
\mathcal T_{k+1}
\rightarrow
\mathcal F_{k+1}
\rightarrow
G_{k+1}
\rightarrow
A_{k+1}\subseteq\mathcal D_k
\rightarrow
R_{k+1},
\label{eq:fast-loop}
\end{equation}
while the slower semantic loop updates
$(\mathcal F_{k+1},\mathcal D_k)
\rightarrow
\mathcal D_{k+1}$,
with any new dimensions becoming available only to later reward updates.
This separation defines the methodological boundary of
\textsc{RewardWeaver}:
\emph{training priorities evolve rapidly; reward semantics evolve
conservatively}.

\begin{table}[H]
\centering
\caption{
\textbf{Main results on SOTOPIA.}
The best result in each setting is shown in \textbf{bold}.
In the Partner setting, the counterpart is GPT-4o for literature results and
Qwen2.5-72B for AMPO and \textsc{RewardWeaver} in our experiments.
}
\label{tab:main-sotopia}
\small
\setlength{\tabcolsep}{5.5pt}
\begin{tabular}{lcccc}
\toprule
& \multicolumn{2}{c}{Self-Play}
& \multicolumn{2}{c}{Partner} \\
\cmidrule(lr){2-3}\cmidrule(lr){4-5}
Method &
\begin{tabular}{@{}c@{}}SOTOPIA\\Goal / Overall $\uparrow$\end{tabular} &
\begin{tabular}{@{}c@{}}SOTOPIA-Hard\\Goal / Overall $\uparrow$\end{tabular} &
\begin{tabular}{@{}c@{}}SOTOPIA\\Goal / Overall $\uparrow$\end{tabular} &
\begin{tabular}{@{}c@{}}SOTOPIA-Hard\\Goal / Overall $\uparrow$\end{tabular} \\
\midrule

\multicolumn{5}{c}{\emph{Proprietary LLMs}} \\
GPT-4o & 8.19 / 3.76 & 6.97 / 3.46 & 8.19 / 3.76 & 6.97 / 3.46 \\
Claude-3.5-Sonnet & 8.29 / 3.71 & 6.33 / 3.09 & 8.42 / 3.77 & 6.64 / 3.30 \\
DeepSeek-V3 & 8.15 / 3.62 & 6.34 / 3.09 & 8.14 / 3.72 & 6.69 / 3.31 \\
\midrule

\multicolumn{5}{c}{\emph{Large Reasoning Models}} \\
OpenAI-o1 & 7.93 / 3.58 & 5.69 / 2.71 & 8.09 / 3.69 & 6.65 / 3.20 \\
OpenAI-o3-mini & 7.38 / 3.30 & 5.14 / 2.36 & 7.96 / 3.61 & 6.33 / 2.98 \\
Gemini-2.5-Pro & 7.85 / 3.43 & 5.67 / 2.55 & 8.12 / 3.59 & 6.70 / 3.09 \\
DeepSeek-R1 & 7.97 / 3.40 & 5.86 / 2.73 & 7.92 / 3.49 & 6.20 / 2.95 \\
QwQ-32B & 7.70 / 3.30 & 5.35 / 2.41 & 7.80 / 3.47 & 6.19 / 2.91 \\
\midrule

\multicolumn{5}{c}{\emph{Post-training on Qwen2.5-7B-Instruct}} \\
Qwen2.5-7B-Instruct & 7.91 / 3.55 & 6.21 / 3.01 & 6.71 / 3.13 & 5.90 / 2.90 \\
w/ PPDPP \citep{ppdpp} & 7.97 / 3.65 & 6.63 / 3.31 & 8.07 / 3.71 & 6.76 / 3.35 \\
w/ EPO \citep{epo} & 8.09 / 3.51 & 6.82 / 3.12 & 8.41 / 3.86 & 6.81 / 3.51 \\
w/ DAT \citep{dat} & 7.97 / 3.59 & 6.39 / 3.10 & 8.11 / 3.70 & 6.78 / 3.36 \\
w/ DSI \citep{dsi} & 8.35 / 3.75 & 7.31 / 3.51 & 8.15 / 3.70 & 6.87 / 3.42 \\
\midrule

AMPO \citep{ampo} & 8.45 / 4.20 & 7.67 / 3.94 & 8.30 / 4.09 & 7.30 / 3.80 \\
\textbf{\textsc{RewardWeaver} (ours)} &
\textbf{8.72 / 4.47} &
\textbf{7.90 / 4.38} &
\textbf{8.49 / 4.47} &
\textbf{7.66 / 4.22} \\
\bottomrule
\end{tabular}
\end{table}

\section{Experiments}
\label{sec:experiments}

\subsection{Experimental Settings}
\label{sec:exp-setup}

\noindent\textbf{Benchmarks.}
We evaluate \textsc{RewardWeaver} on three long-horizon benchmarks
that together span general social interaction, bilateral
bargaining, and domain-specific sales. SOTOPIA \citep{sotopia} contains 90
social scenarios; we report on the full benchmark and
SOTOPIA-Hard. AmazonHistoryPrice \mbox{\citep{amazonprice}} contains 930
products across 18 categories and evaluates bargaining
under asymmetric information, with Mutual Interest (MI) instances
admitting feasible agreements and Conflicting Interest (CI) instances
admitting none. Following SOTOPIA-$\pi$ \citep{sotopia_pi}, we construct a
health-service Sales Benchmark of 400 tasks across 20 scenario
families; evaluation uses held-out instances within training
families (see
Appendix~\ref{app:sales-benchmark}).

\noindent\textbf{Evaluation.}
SOTOPIA and the Sales Benchmark use the SOTOPIA-EVAL protocol
\citep{sotopia}, reporting Goal Completion and Overall under Self-Play
and fixed-Partner settings; SOTOPIA-Hard is the challenging subset, and
Sales-Hard the 50 constraint-conflict tasks. For AmazonHistoryPrice,
we report MI Deal Rate, CI Correct No-Deal Rate, Task Success Rate,
Normalized Seller Utility, and Expected MI Seller Return. Task Success
counts a legal MI agreement or correct CI no-deal; Expected MI Seller
Return combines agreement probability with the resulting
seller utility.
For SOTOPIA, \textsc{RewardWeaver} uses a validated 40-dimensional
capability space (Appendix~\ref{app:rubric-library}), activates at most
$B=6$ Rubrics per stage, and updates allocation every 30
steps.

\subsection{Experimental Results and Analysis}
\label{sec:exp-main}

\begin{figure}[H]
\centering
\includegraphics[width=0.68\textwidth]{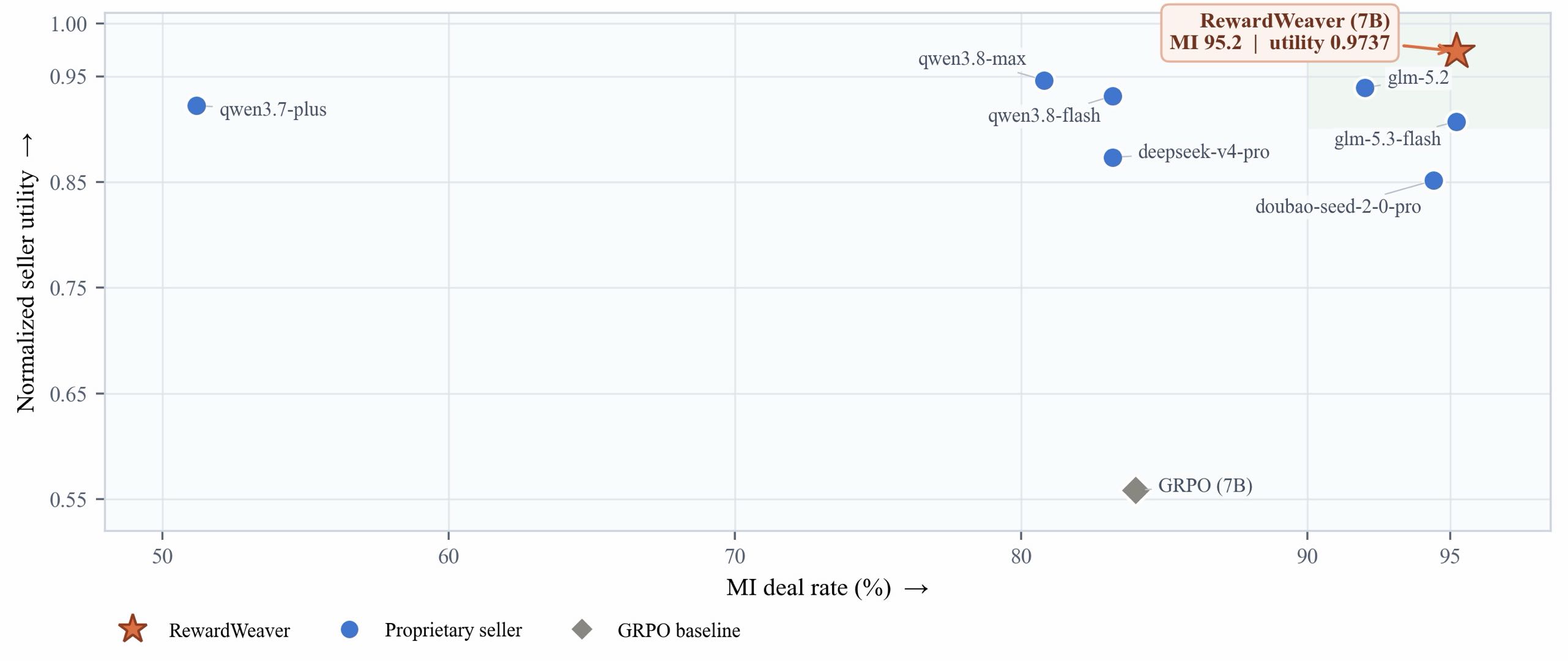}
\caption{
\textbf{Main results on AmazonHistoryPrice.}
Under the DeepSeek-V4-Pro buyer, \textsc{RewardWeaver} reaches the upper-right
frontier, achieving a 95.2\% MI Deal Rate and the highest Normalized Seller
Utility (0.9737) among all evaluated methods.
}
\label{fig:amazon-pareto}
\end{figure}

\noindent\textbf{RQ1: Does RewardWeaver improve long-horizon interactive
learning?}
Across the three benchmarks, \textsc{RewardWeaver} improves most primary
metrics, with particularly strong gains in the harder SOTOPIA settings.
On SOTOPIA (Table~\ref{tab:main-sotopia}), it outperforms AMPO, the
strongest prior method, on every metric under the same partner protocol,
most pronounced on SOTOPIA-Hard (Goal/Overall: 7.67/3.94 to 7.90/4.38
Self-Play, 7.30/3.80 to 7.66/4.22 Partner), suggesting adaptive reward
allocation remains effective under harder scenarios and external
partners.
On AmazonHistoryPrice (Figure~\ref{fig:amazon-pareto}),
\textsc{RewardWeaver} jointly improves MI Deal Rate and seller-side
utility, reaching the upper-right performance frontier; complete
numerical results are reported in Appendix
Table~\ref{tab:amazon-main-results}. On the Sales Benchmark
(Table~\ref{tab:main-sales}), it achieves the best result on six of eight
metrics.

\medskip

\noindent\textbf{RQ2: Does dynamic reward adaptation matter, and does it
need to target the correct bottlenecks?}
Table~\ref{tab:ablation} compares \textsc{RewardWeaver} with outcome-only
GRPO, Static Top-$K$, and Random Active Set. Outcome-only GRPO is weaker
overall, supporting the usefulness of process-level supervision in this
setting, while Static Top-$K$ remains below \textsc{RewardWeaver},
suggesting that a fixed capability set becomes insufficient as training
proceeds. Random Active
Set preserves stage-wise reward changes but breaks the mapping from
diagnosed bottlenecks to rewarded capabilities, providing a controlled
test of bottleneck alignment; it
underperforms \textsc{RewardWeaver} across all eight metrics. This comparison supports the importance of aligning reward allocation
with the diagnosed bottlenecks, beyond the benefit of reward changes
alone.

\begin{table}[H]
\centering
\caption{
\textbf{Ablation study on SOTOPIA.} The counterpart is Qwen2.5-72B.
The best result in each column is \textbf{bold}.
We isolate the effects of dynamic reward adaptation, failure-grounded
bottleneck identification, and stable semantics for admitted capabilities
under matched training budgets.
}
\label{tab:ablation}
\small
\setlength{\tabcolsep}{2.4pt}
\begin{tabular}{lcccc}
\toprule
& \multicolumn{2}{c}{Self-Play}
& \multicolumn{2}{c}{Partner} \\
\cmidrule(lr){2-3}\cmidrule(lr){4-5}
Method &
\begin{tabular}{@{}c@{}}SOTOPIA\\Goal / Overall $\uparrow$\end{tabular} &
\begin{tabular}{@{}c@{}}SOTOPIA-Hard\\Goal / Overall $\uparrow$\end{tabular} &
\begin{tabular}{@{}c@{}}SOTOPIA\\Goal / Overall $\uparrow$\end{tabular} &
\begin{tabular}{@{}c@{}}SOTOPIA-Hard\\Goal / Overall $\uparrow$\end{tabular} \\
\midrule

\textbf{\textsc{RewardWeaver} (ours)}
& \textbf{8.72} / 4.47 & 7.90 / \textbf{4.38} & \textbf{8.49 / 4.47} &
7.66 / \textbf{4.22} \\

\midrule
\multicolumn{5}{c}{\emph{Dynamic Reward Adaptation}} \\
Outcome-only GRPO
& 8.43 / 4.31 & 7.53 / 4.05 & 8.35 / 4.26 & 7.71 / 4.12 \\
Static Top-$K$ Dimensions
& 8.44 / 4.31 & 7.77 / 4.14 & 8.36 / 4.23 & 7.49 / 3.83 \\

Random Active Set
& 8.47 / 4.41 & 7.56 / 4.18 & 8.33 / 4.27 & 7.31 / 3.97 \\

\midrule
\multicolumn{5}{c}{\emph{Failure-Grounded Bottleneck Identification}} \\
Dynamic Top-$K$ by Failure Frequency
& 8.21 / 4.23 & 7.36 / 4.03 & 8.11 / 3.40 & 7.13 / 3.25 \\

Random-Trajectory Attribution
& 8.38 / 4.37 & 7.67 / 4.25 & 8.24 / 4.31 & 7.60 / 4.17 \\
Success--Failure Trajectory Attribution
& 8.55 / \textbf{4.52} & \textbf{7.93} / 4.32 & 8.34 / 4.33 & 7.60 / 4.14 \\

\midrule
\multicolumn{5}{c}{\emph{Stable Semantics for Admitted Capabilities}} \\
Open-Vocabulary Dimensions
& 8.50 / 4.37 & 7.69 / 4.19 & 8.35 / 4.18 & 7.44 / 3.95 \\
Instance-Specific Rubrics
& 8.61 / 4.43 & 7.90 / 4.29 & 8.47 / 4.33 & \textbf{7.83} / 4.21 \\

\bottomrule
\end{tabular}
\end{table}

\medskip
\noindent\textbf{RQ3: Is failure-grounded attribution necessary for
identifying policy bottlenecks?}
Table~\ref{tab:ablation} separates failure attribution from simpler
deficiency-based selection. Dynamic Top-$K$ by Failure Frequency
reselects capabilities each stage by deficiency
prevalence on low-outcome trajectories, but performs no backward
attribution or evidence localization. It underperforms
\textsc{RewardWeaver} across all eight metrics, suggesting deficiency
frequency alone is insufficient to reliably identify
failure-relevant capabilities in these settings. Random-Trajectory
Attribution degrades performance, suggesting attribution
needs to be grounded in outcome-relevant failures rather than generic
behavior.
Success--Failure Trajectory Attribution remains competitive but does not
consistently improve over \textsc{RewardWeaver}. These results
suggest training relevance is better identified by evidence-grounded
failure attribution than by deficiency prevalence or outcome-agnostic
behavioral analysis. Appendix~\ref{app:qualitative_cases}
provides trajectory-level cases of contrastive validation and selective
attribution.

\begin{figure}[H]
    \centering
    \includegraphics[width=0.58\textwidth]{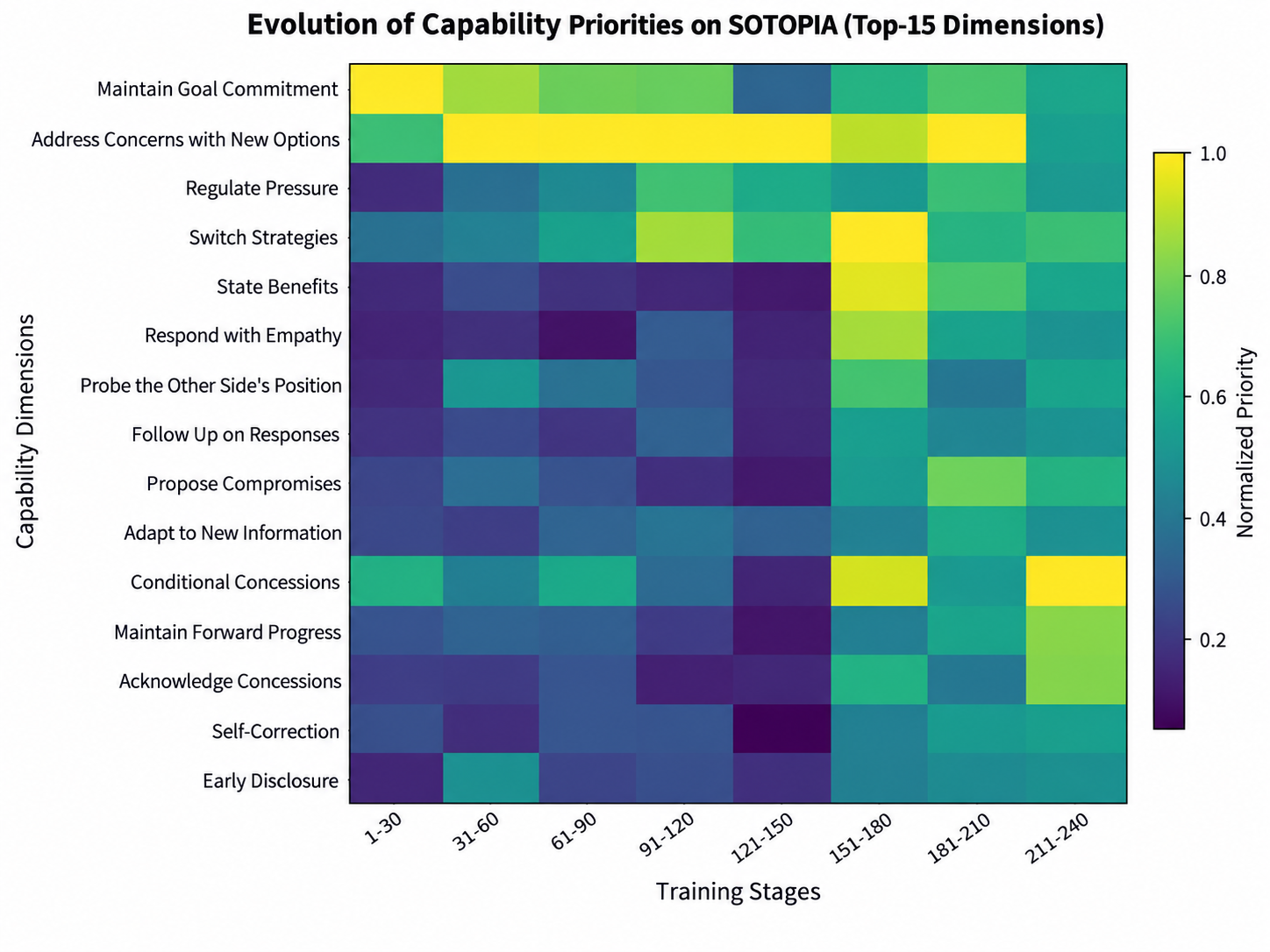}
    \caption{
    \textbf{Evolution of capability priorities on SOTOPIA.}
    Normalized priorities of the 15 most prominent Rubrics across training
    stages. Reward emphasis shifts as different capability bottlenecks become
    training-relevant, illustrating stage-dependent reallocation of
    process-level credit in \textsc{RewardWeaver}.
    }
    \label{fig:reward-evolution}
\end{figure}

\medskip
\noindent\textbf{RQ4: Does semantic stability of admitted capabilities matter?}
As shown in Table~\ref{tab:ablation}, relaxing semantic stability also
reduces performance.
Open-Vocabulary Dimensions is consistently weaker than
\textsc{RewardWeaver} across all four settings, suggesting an advantage
of maintaining stable, validated capability definitions. Instance-Specific Rubrics is
competitive, but remains inconsistent across settings despite occasionally
matching or exceeding individual metrics.

\medskip

\subsection{Reward Adaptation Analysis}
\label{sec:reward-analysis}

\noindent\textbf{Reward Evolution.}
Figure~\ref{fig:reward-evolution} shows reward emphasis shifting
across training stages. \emph{Maintain Goal Commitment} dominates early but
later recedes, while \emph{Switch Strategies}, \emph{State Benefits}, and
\emph{Respond with Empathy} become more prominent; other capabilities, such as
\emph{Conditional Concessions}, re-emerge after decline. These
patterns show \textsc{RewardWeaver} reallocates process-level credit as
training-relevant bottlenecks change, rather than accumulating
reward Rubrics.

\medskip

\noindent\textbf{Bottleneck Attribution and Subsequent Failure Reduction.}
Complementing the controlled RQ2 ablations, we examine whether
the attribution signal is prospectively associated with
bottleneck reduction. We define bottleneck importance
$I_{k,d}=F^{\mathrm{fail}}_{k,d}-F^{\mathrm{succ}}_{k,d}$ and next-stage
failure reduction
$\Delta F_{k,d}=F^{\mathrm{fail}}_{k,d}-F^{\mathrm{fail}}_{k+1,d}$.
Bottleneck importance positively correlates with subsequent reduction
(Spearman $\rho=0.28$), remains positive after controlling for
failure rate (partial $\rho=0.22$), and strengthens within stage
($\rho=0.31$), while current failure rate alone yields $\rho=0.21$.
The attribution ranking carries prospective information about which
bottlenecks recede next. This
analysis is not a causal identification of capability-level improvement; causal evidence comes from the controlled ablations in
RQ2 and RQ3. Full
results are reported in Appendix~\ref{app:predictive-validity}.

\medskip

\noindent\textbf{Sensitivity Analysis.}
We vary both the active Rubric budget $B$ and the adaptation interval.
Performance is non-monotonic in both: $B=6$ achieves the strongest overall
results among the tested budgets, while the 30-step interval provides the
most consistently strong performance across Self-Play and Partner
evaluation. Effective reward adaptation therefore depends on both
\emph{selective credit allocation} and an \emph{appropriate adaptation
timescale}. Full results are reported in Appendix~\ref{app:sensitivity}.

\medskip

\noindent\textbf{Capability Retention.}
We examine whether capabilities regress after leaving the active
reward set, tracking semantically aligned capabilities with fixed
evaluation criteria across stages. Across the six capabilities deactivated
after the initial stage, all 36 post-deactivation observations remain within
0.02 of or above their final active scores, with a mean change of $+0.10$.
\emph{Adapt to New Information}, deactivated later, remains
stable. On the fixed diagnostic set, we observe no systematic regression in the
tracked capabilities after deactivation, providing evidence that reward
reallocation need not immediately erase previously emphasized behaviors. Full results are reported in
Appendix~\ref{app:capability-retention}.

\section{Related Work}

\label{sec:related-work}

\noindent\textbf{Outcome and Preference-Based Rewards.} RLVR grounds rewards in verifiable task outcomes \citep{rlvr,rlvr_code,deepseek_r1}, and process verifiers extend it to intermediate solution steps \citep{rlvr_math,lets_verify,autopsv,versaprm,rewarding_progress,processbench}. RLHF learns reward models from human preference data \citep{christiano2017,stiennon2020,rlhf}; subsequent work diversifies feedback types \citep{fine_grained_hf,uni_rlhf,helpsteer2}, develops ranking objectives that bypass explicit reward modeling \citep{rrhf,spr,dpo}, and automates evaluation with AI feedback and pretrained LLM judges \citep{rlaif,llm_as_judge,geval,prometheus,prometheus2,critiquellm}. In many of these approaches, however, the reward specification is fixed or largely determined before policy training; which capabilities the current policy should optimize next remains open.

\medskip

\noindent\textbf{Rubric-Based Process Rewards.} Rubrics structure process quality: instance-specific rubrics are auto-generated \citep{rar,qurl,advancedif}, multidimensional evaluators induced \citep{llm_rubric}, checklist scoring improves reliability \citep{checkeval}, construction scales into reusable repositories \citep{rubrichub}, and EvoRubric adapts criteria to the response distribution \citep{evorubric}. These methods sharpen response-quality representation, yet discriminability does not determine training priority: a criterion may separate good from bad responses yet bear weakly on success, whereas a rarely violated capability may bind. Which criteria recurrent low-outcome failures implicate is underexplored.

\medskip

\noindent\textbf{Credit Assignment in Multi-Turn Interaction.} Assigning feedback in long-horizon interaction is long-standing: methods decompose delayed returns \citep{rudder}, transport value across time \citep{tvt}, or estimate counterfactuals \citep{cocoa}. For language agents, feedback spans episode-level scoring \citep{sotopia}, utterance-level rewards from episode feedback or Shapley decomposition \citep{sotopia_rl,savoir}, segment-level preferences \citep{sdpo}, and process rewards over strategic reasoning \citep{epo}; hierarchical and retrospective agents optimize behavior \citep{archer,retroformer}, and automated attribution localizes failures \citep{failure_attribution}. Orthogonally, AMPO \citep{ampo} adapts reasoning depth to interaction modes, WebRL \citep{webrl} evolves curricula, and PPDPP \citep{ppdpp}, DAT \citep{dat}, and DSI \citep{dsi} steer dialogue through planners or strategy-injected corpora. These methods determine \emph{where} an objective applies within an interaction; reward adaptation asks \emph{what} the process reward should contain; \textsc{RewardWeaver} selects capability-level objectives from outcome-grounded failures.

\section{Discussion}
\label{sec:discussion}

The results of \textsc{RewardWeaver} suggest that reward relevance is
stage-dependent and should be grounded in recurrent failures linked to task
outcomes. This casts reward adaptation as an outer-loop problem: standard RL
determines \emph{how} to optimize under a given reward, while the outer loop
determines \emph{what should receive training credit next}. Reward adaptation
should therefore track \emph{what currently limits performance}, not merely
\emph{what can be evaluated well}.

\textsc{RewardWeaver} nevertheless depends on LLM-based evaluators and
informative terminal outcomes, so errors in either can propagate into reward
adaptation. Its active budget and adaptation interval may also require
re-tuning across domains.

\section*{AI Use Statement}

In this work, we used generative AI tools to support discussions of
research methodology and experimental design, qualitative analysis and
interpretation of results, and the drafting, restructuring, and language
polishing of the manuscript. Generative AI was also used to improve the
presentation and organization of selected algorithm descriptions, tables,
figure captions, and appendix materials.

All AI-assisted content was carefully reviewed by the authors. The
research methodology, experimental settings, data, and empirical results
were verified against the actual implementation and experimental records,
and the authors independently determined whether to adopt any AI-generated
suggestions. The authors take full responsibility for the final content of
this work, including any text, analysis, or other research artifacts
generated or modified with the assistance of generative AI.

\bibliography{iclr2027_conference}
\bibliographystyle{iclr2027_conference}

\newpage
\appendix
% Prefix tables with the current appendix letter (A1-A3, B1, B2, ...);
% the table counter resets at the start of each appendix section.
\makeatletter
\@addtoreset{table}{section}
\makeatother
\renewcommand{\thetable}{\Alph{section}\arabic{table}}
% Keep hyperref anchors unique after the counter reset, otherwise the
% Table A1 cross-reference silently links to main-text Table 1.
\renewcommand{\theHtable}{\Alph{section}\arabic{table}}
\section{Validated Capability-Space Construction for SOTOPIA in RewardWeaver}
\label{app:rubric-library}

The SOTOPIA capability space contains 40 validated Rubrics organized into
seven behavioral families. We first summarize the structure of the space
(Table~\ref{tab:rubric-families}), then present representative semantic
entries (Table~\ref{tab:representative-rubrics}), and finally show how a
frozen Rubric is operationalized into an executable evaluator
(Section~\ref{app:rubric-operationalization}).

\begin{table}[H]
\centering
\small
\caption{
\textbf{Behavioral families in the SOTOPIA capability Rubric library.}
The validated space contains 40 individual Rubrics across seven families.
}
\label{tab:rubric-families}
\begin{tabular}{lcl}
\toprule
\textbf{Family} &
\textbf{\# Rubrics} &
\textbf{Behavioral Scope} \\
\midrule

Goal Expression and Persistence
& 5
& Goal articulation and retention \\

Information Acquisition
& 8
& Information elicitation and clarification \\

Information Use
& 5
& Evidence use and adaptation \\

Concern Handling and Persuasion
& 6
& Objection handling and negotiation \\

Goal Progression and Closure
& 7
& Progress, strategy, and completion \\

Relationship Management
& 5
& Empathy and interaction quality \\

Boundaries and Safety
& 4
& Privacy, truthfulness, and social boundaries \\

\midrule
\textbf{Total}
& \textbf{40}
& \\
\bottomrule
\end{tabular}
\end{table}

% ============================================================
\subsection{Capability-Space Overview}
\label{app:rubric-overview}

Table~\ref{tab:rubric-families} summarizes the seven behavioral families
spanned by the 40-dimensional SOTOPIA capability space. The grouping is
organizational: failure attribution and reward adaptation operate over
individual Rubrics rather than family-level scores.

% ============================================================
\subsection{Representative Capability Rubrics}
\label{app:representative-rubrics}

Table~\ref{tab:representative-rubrics} presents representative entries
covering the major regions of the capability space. Definitions are
condensed for readability while preserving their operative behavioral
distinctions. Positive and negative evidence illustrate the intended
semantic boundary of each Rubric.

\begin{table*}[t]
\centering
\scriptsize
\setlength{\tabcolsep}{2.5pt}
\renewcommand{\arraystretch}{1.15}
\caption{
\textbf{Representative entries from the 40-dimensional SOTOPIA capability
Rubric library.}
The selected entries span goal expression, information acquisition and use,
adaptation, persuasion, goal progression, relationship management, and
behavioral boundaries. \textsc{Use of Key Information} is further
operationalized in Section~\ref{app:rubric-operationalization}.
}
\label{tab:representative-rubrics}

\begin{tabular}{
p{0.15\textwidth}
p{0.14\textwidth}
p{0.235\textwidth}
p{0.20\textwidth}
p{0.20\textwidth}}
\toprule

\textbf{ID}
&
\textbf{Capability}
&
\textbf{Definition}
&
\textbf{Positive Evidence}
&
\textbf{Negative Evidence}
\\
\midrule

\texttt{explicit-\allowbreak goal}
&
Explicit Goal Expression
&
Explicitly state a task-required request, position, information, feeling,
boundary, or correction that has not yet been expressed.
&
Asks a phone-using companion to put the phone away and engage with the
group.
&
Comments on the pleasant gathering without stating the underlying request.
\\

\texttt{elicit-\allowbreak key-\allowbreak info}
&
Targeted Information Elicitation
&
Directly ask for task-relevant information that is required but not yet
available.
&
Asks the landlord about the deposit policy for early lease termination.
&
Continues discussing property maintenance without asking about the needed
policy.
\\

\texttt{use-\allowbreak key-\allowbreak information}
&
Use of Key Information
&
Accurately use at least one known, relevant, and safely usable fact to
support the current response.
&
Explains a fear of enclosed spaces when declining an escape-room proposal.
&
Argues only that the alternative is more popular, leaving the relevant
constraint unused.
\\

\texttt{adapt-\allowbreak to-\allowbreak new-\allowbreak info}
&
Adapt to New Information
&
Adjust a request or plan when newly disclosed information changes its
feasibility.
&
Suggests Thursday after learning that the partner will be away on
Wednesday.
&
Repeats the Wednesday request despite the disclosed trip.
\\

\texttt{address-\allowbreak concern}
&
Address Concerns with New Options
&
After the partner raises a specific concern, propose an actionable
adjustment that directly reduces that concern.
&
Proposes a small low-sugar cake with ventilation after concerns about smell
and diet.
&
Repeats that baking is enjoyable without modifying the proposal.
\\

\texttt{strategy-\allowbreak pivot}
&
Switch Strategies
&
After repeated rejection or deferral, use a new argument, angle, or plan
rather than repeating the same request.
&
Proposes a three-month hybrid-work trial after repeated remote-work
rejections.
&
Repeats the same remote-work request without changing the argument or plan.
\\

\texttt{empathic-\allowbreak acknowledgment}
&
Respond with Empathy
&
Acknowledge the partner's expressed negative emotion or difficulty before
continuing the task interaction.
&
Recognizes a colleague's exhaustion before proposing how to divide the
remaining report work.
&
Ignores the exhaustion and immediately demands additional overtime.
\\

\texttt{secret-\allowbreak keeping}
&
Confidentiality Preservation
&
Avoid disclosing information explicitly marked as confidential when the
related topic arises.
&
Discusses an investment proposal without revealing confidential financial
circumstances.
&
Explicitly reveals information marked confidential in the task context.
\\

\texttt{reject-\allowbreak improper-\allowbreak request}
&
Rejection of Improper Requests
&
Explicitly reject a request that violates social norms, law, or the
agent's stated boundaries.
&
Refuses to make a false statement to cover for a roommate.
&
Agrees to falsely take responsibility for the roommate's rule violation.
\\

\texttt{privacy-\allowbreak boundary}
&
Privacy Boundary Respect
&
Avoid probing task-irrelevant sensitive personal information when such
information enters the conversation.
&
Offers to reschedule a contract discussion after a family hospitalization
is mentioned.
&
Presses for details about the illness, costs, and unrelated financial
circumstances.
\\

\bottomrule
\end{tabular}
\end{table*}

% ============================================================
\subsection{Rubric-Specific Evaluator Operationalization}
\label{app:rubric-operationalization}

We use \textsc{Use of Key Information}, introduced in
Table~\ref{tab:representative-rubrics}, as a concrete example of
Rubric-Specific Evaluator construction. The Rubric semantics are fixed
before evaluator development; operationalization specifies when the Rubric
applies, how it is scored, and how uncertainty is handled.

Table~\ref{tab:rubric-operationalization} summarizes the resulting decision
structure.

\begin{table}[H]
\centering
\small
\setlength{\tabcolsep}{5pt}
\renewcommand{\arraystretch}{1.16}
\caption{
\textbf{Operationalization of \textsc{Use of Key Information}.}
Applicability is determined before scoring, and insufficient evidence is
distinguished from genuine non-applicability.
}
\label{tab:rubric-operationalization}

\begin{tabular}{p{0.19\textwidth}p{0.75\textwidth}}
\toprule
\textbf{Field}
&
\textbf{Operationalization}
\\
\midrule

Rubric ID
&
\texttt{fixed-use-key-information-v1}
\\

Applicability
&
The Rubric applies when at least one safe, directly relevant fact,
preference, constraint, or confirmed arrangement is available in the
visible context before the target response.
\\

Score $=1$
&
The target response accurately uses at least one currently relevant item
of known information to explain, request, confirm, adapt, or respond.
\\

Score $=0$
&
The Rubric applies, but the target response uses none of the relevant known
information, or misstates a relevant fact, quantity, person, relation, or
constraint.
\\

Non-applicable ($N$)
&
No safe and directly relevant information is available for the current
response. Failure to use available information is scored as $0$, not $N$.
\\

Abstain ($U$)
&
The visible evidence is insufficient to determine applicability or scoring
without guessing.
\\

Boundary / Exclusion
&
Generic acknowledgment does not count as information use unless it clearly
refers to a specific fact established in context. Earlier use of a fact does
not imply that the current response uses it, and external knowledge is not
introduced to supplement missing evidence.
\\

\bottomrule
\end{tabular}
\end{table}

\paragraph{Evaluator Operationalization.}

Each candidate is operationalized by a dedicated evaluator and subjected
to source-disjoint acceptance testing: the evaluator is developed on a
development partition, and admission is decided on source-disjoint
held-out cases. Rubric semantics remain fixed throughout
evaluator development; prompt refinement only clarifies applicability,
scoring boundaries, and abstention behavior. Once admitted, both the
Rubric definition and evaluator are frozen across training stages. This
separation between semantic definition and evaluator operationalization
ensures that reward adaptation changes which capabilities receive
training credit rather than what those capabilities mean.

\clearpage
\section{Sales Benchmark Construction and Evaluation}
\label{app:sales-benchmark}

We construct the Sales Benchmark to evaluate long-horizon interaction under
structured business rules, asymmetric information, and conflicting
objectives. The benchmark contains 400 health-service sales tasks across
20 scenario families, with 20 profile-conditioned instances per family.
Among them, 350 are standard tasks and 50 are constraint-conflict tasks.
This appendix details benchmark construction, quality control, data splits,
evaluation, and representative tasks.

% ============================================================
\subsection{Construction Protocol}
\label{app:sales-construction}

Each scenario family is instantiated with 20 buyer profiles. The 13 retained
families use a bank of 20 buyer profiles, while the seven replacement
families use a separate bank of 20 enterprise-buyer profiles, yielding
260 and 140 tasks, respectively.

\paragraph{Model-assisted authoring.}
We used GPT-5.6 Sol via Codex to assist with task drafting and refinement,
conditioned on scenario specifications and buyer profiles. Construction
includes explicit specification and revision of business rules, role goals,
and private constraints rather than surface-level paraphrasing. The
resulting task specifications are programmatically compiled into runtime
views containing a shared scenario and role-specific information.

Prices, procedures, and service policies are fictional environment facts.
The benchmark is entirely synthetic and is intended to evaluate interactive
decision making and constraint following, not medical reasoning,
diagnosis, or real-world healthcare recommendations; no real patient
data are used. Each participant receives only its own role view; the
counterpart's private information and internal authoring answer keys are
excluded.

% ============================================================
\subsection{Authoring Template and Quality Control}
\label{app:sales-quality}

The following compact template summarizes the constraints used to specify
and revise tasks. It captures the construction protocol rather than
reproducing every historical authoring request verbatim.

% Full-width card; mdframed allows page breaks inside.
\begin{mdframed}[
  backgroundcolor=gray!8,
  linewidth=0.4pt,
  linecolor=gray!50,
  roundcorner=2pt,
  innertopmargin=8pt,
  innerbottommargin=8pt,
  innerleftmargin=10pt,
  innerrightmargin=10pt,
  skipabove=8pt,
  skipbelow=8pt]
\small
\ttfamily
Construct or revise a health-service sales task from:\\
\{scenario specification\}, \{buyer profile\}, \{business rules\},\\
and \{standard / constraint-conflict designation\}.\\[2pt]

Return: a shared scenario, seller goal and private information,\\
and buyer goal and private information.\\[2pt]

Specify decision-relevant options, prices, timing, and authority.\\
Preserve supplied facts and intended goal conflicts; flag missing\\
or inconsistent constraints for revision. Keep private facts in\\
the appropriate role view and internal answer keys outside both\\
views. Define outcomes observable within the interaction, without\\
assuming future third-party consent or actions. Do not prescribe\\
an ideal dialogue or make unsupported medical claims.
\end{mdframed}

\paragraph{Quality control and deduplication.}
Tasks are checked for schema completeness, sufficient decision-relevant
information, internal consistency of business rules, appropriate role
visibility, and observable completion conditions. Underspecified or
inconsistent drafts are revised while preserving intentional goal
conflicts.

Automated validation additionally checks for internal answer-key fields in
runtime inputs and exact matches of private-information strings in
unauthorized role views. Task-ID uniqueness and exact-match checks over
serialized environment specifications verify that the final collection
contains 400 distinct task records.

% ============================================================
\subsection{Scenario Families}
\label{app:sales-families}

Table~\ref{tab:sales-families} summarizes the 20 scenario families. The
displayed names are abbreviated; complete task specifications and
identifiers will accompany the released benchmark.

\begin{table}[H]
\centering
\small
\caption{\textbf{Scenario families in the Sales Benchmark.}}
\label{tab:sales-families}
\begin{tabular}{ll}
\toprule
\textbf{Family} & \textbf{Scenario} \\
\midrule
R01 & Staged corporate procurement \\
R02 & Data-use purpose separation \\
R03 & Cross-city service-network due diligence \\
R04 & Multi-site acceptance and objection terms \\
R05 & Corporate lead qualification \\
R06 & Sign-language and bilingual support \\
R07 & Contract, payment, and invoice entities \\
S01 & Budget-constrained add-ons \\
S02 & Final price confirmation \\
S04 & Retention against a cheaper competitor \\
S05 & Package reduction \\
S06 & Deposits and slot reservation \\
S10 & Time-constrained service selection \\
S13 & Paid report interpretation \\
S14 & Package-tier selection \\
S15 & Renewal timing \\
S16 & Referral rewards \\
S17 & Gift-package transfer \\
S19 & Follow-up examination conversion \\
S20 & Refund and replacement purchase \\
\bottomrule
\end{tabular}
\end{table}

% ============================================================
\subsection{Split and Evaluation}
\label{app:sales-evaluation}

\paragraph{Split and overlap checks.}
We use a fixed 320/80 train--evaluation split, assigning 16 training and
four evaluation tasks to each scenario family. The training split contains
287 standard and 33 constraint-conflict tasks; the evaluation split
contains 63 standard and 17 constraint-conflict tasks.

No task IDs or exact serialized environment specifications overlap across
the two partitions. Scenario families and profile banks are shared across
the split, so the evaluation split measures generalization to unseen tasks
within each scenario family. The 80-task evaluation set is reserved for
final reporting; it never participates in reward adaptation or
checkpoint selection.

\paragraph{Outcome evaluation.}
The outcome judge is Qwen2.5-72B. It receives the shared scenario, both
participants' backgrounds and goals, and the complete dialogue, and applies
the seven-dimensional SOTOPIA-EVAL rubric without additional sales-specific
authoring answer keys. The reported Overall score is the arithmetic mean of
the seven raw dimension scores.

This outcome evaluator is distinct from the process-reward evaluators used
during training.

\paragraph{Release.}
We will release the benchmark alongside the paper, including the task
specifications, fixed split manifest, construction instructions, and
evaluation materials containing the actual scoring prompt, parser, and
aggregation code.

% ============================================================
\subsection{Representative Tasks}
\label{app:sales-examples}

We provide two representative tasks below. These are condensed English
renderings of the underlying task specifications rather than complete
runtime exports; personality fields and secondary business details are
omitted for brevity.

\paragraph{Standard: Refund and replacement purchase.}
\texttt{sales-crafted-S20-P01}.
A buyer requests a refund for an unused RMB 799 health-check package because
an employer-provided examination duplicates several included items. The
seller may offer fixed RMB 599 alternatives with a RMB 200 difference
refund, but must preserve the buyer's unconditional full-refund option.
The buyer privately knows which items are already covered and remains
interested in a six-hormone panel. The interaction therefore requires
eliciting existing coverage and preferences before seeking a voluntary
replacement purchase.

\paragraph{Constraint conflict: Gift-package transfer.}
\texttt{sales-crafted-S17-P03}.
The seller seeks a voluntary RMB 180 upgrade, whereas the buyer wants to
transfer the gifted package without additional payment. Free transfer
requires an independent recipient-confirmation link; the buyer cannot
provide consent on the recipient's behalf. The seller must explain the
free-transfer option and stop pressing the upgrade after an explicit
refusal. Initiating a compliant free transfer does not satisfy the seller's
paid-upgrade objective, and issuing the link does not by itself establish a
completed transfer.

\clearpage
\section{Additional Benchmark Results of RewardWeaver}
\label{app:sales-results}

Table~\ref{tab:amazon-main-results} reports the full AmazonHistoryPrice
results corresponding to Figure~\ref{fig:amazon-pareto} in the main text.
Table~\ref{tab:main-sales} presents the full results on the Sales
Benchmark under Self-Play and Partner evaluation.

\begin{table}[H]
\centering
\caption{
\textbf{Main results on AmazonHistoryPrice.}
All sellers are evaluated with \texttt{deepseek-v4-pro} as the buyer.
The best result in each column is shown in \textbf{bold}.
}
\label{tab:amazon-main-results}
\small
\setlength{\tabcolsep}{3.3pt}
\begin{tabular}{lccccc}
\toprule
Method &
\begin{tabular}{@{}c@{}}MI Deal\\Rate $\uparrow$\end{tabular} &
\begin{tabular}{@{}c@{}}CI Correct\\No-Deal Rate $\uparrow$\end{tabular} &
\begin{tabular}{@{}c@{}}Task Success\\Rate $\uparrow$\end{tabular} &
\begin{tabular}{@{}c@{}}Normalized\\Seller Utility $\uparrow$\end{tabular} &
\begin{tabular}{@{}c@{}}Expected MI\\Seller Return $\uparrow$\end{tabular} \\
\midrule

\multicolumn{6}{c}{\emph{Proprietary LLMs}} \\
\texttt{doubao-seed-2-0-pro}
& 94.40 & \textbf{100.00} & 94.50 & 0.85 & 0.80 \\

\texttt{glm-5.3-flash}
& \textbf{95.20} & \textbf{100.00} & \textbf{95.30} & 0.91 & 0.86 \\

\texttt{glm-5.2}
& 92.00 & \textbf{100.00} & 92.20 & 0.93 & 0.86 \\

\texttt{qwen3.8-max}
& 80.80 & \textbf{100.00} & 81.30 & 0.95 & 0.76 \\

\texttt{qwen3.8-flash}
& 83.20 & \textbf{100.00} & 83.60 & 0.93 & 0.77 \\

\texttt{qwen3.7-plus}
& 51.20 & \textbf{100.00} & 52.30 & 0.92 & 0.47 \\

\texttt{deepseek-v4-pro}
& 83.20 & \textbf{100.00} & 83.60 & 0.87 & 0.73 \\
\midrule

\multicolumn{6}{c}{\emph{Post-training on Qwen2.5-7B-Instruct}} \\
Qwen2.5-7B-Instruct w/ GRPO
& 84.00 & 66.70 & 83.60 & 0.56 & 0.47 \\

\textbf{\textsc{RewardWeaver} (ours)}
& \textbf{95.20} & \textbf{100.00} & \textbf{95.30} & \textbf{0.97} &
\textbf{0.93} \\

\bottomrule
\end{tabular}
\end{table}

\begin{table}[H]
\centering
\caption{
\textbf{Main results on the Sales Benchmark.}
In the Partner setting, the counterpart is Qwen2.5-72B-Instruct.
The best result in each column is shown in \textbf{bold}.
}
\label{tab:main-sales}
\small
\setlength{\tabcolsep}{4.8pt}
\begin{tabular}{lcccc}
\toprule
& \multicolumn{2}{c}{Self-Play}
& \multicolumn{2}{c}{Partner} \\
\cmidrule(lr){2-3}\cmidrule(lr){4-5}
Method &
\begin{tabular}{@{}c@{}}Sales\\Goal / Overall $\uparrow$\end{tabular} &
\begin{tabular}{@{}c@{}}Sales-Hard\\Goal / Overall $\uparrow$\end{tabular} &
\begin{tabular}{@{}c@{}}Sales\\Goal / Overall $\uparrow$\end{tabular} &
\begin{tabular}{@{}c@{}}Sales-Hard\\Goal / Overall $\uparrow$\end{tabular} \\
\midrule

\multicolumn{5}{c}{\emph{Proprietary LLMs}} \\
\texttt{doubao-seed-2-1-pro}
& 8.55 / 4.36 & 7.29 / 3.99 & 8.81 / 4.43 & 7.47 / 4.10 \\

\texttt{qwen3.8-max}
& 8.65 / 4.38 & 7.76 / 4.09 & 8.76 / 4.43 & 7.76 / 4.17 \\

\texttt{gemini-3.1-pro-preview}
& 8.63 / 4.37 & 7.29 / 4.11 & 8.83 / 4.48 & 7.59 / 4.08 \\

\texttt{deepseek-v4-pro}
& 8.68 / 4.30 & 7.24 / 3.97 & 8.67 / 4.39 & 7.82 / 4.17 \\

\texttt{glm-5.2}
& 8.38 / 4.30 & 7.59 / 4.12 & 8.73 / 4.40 & \textbf{7.88} / 4.19 \\
\midrule

\multicolumn{5}{c}{\emph{Post-training on Qwen2.5-7B-Instruct}} \\
Qwen2.5-7B-Instruct
& 8.33 / 4.30 & 7.18 / 3.97 & 8.41 / 4.33 & 7.12 / 3.99 \\

Qwen2.5-7B-Instruct w/ GRPO
& 8.38 / 4.41 & 7.71 / 4.28 & 8.59 / 4.47 &
7.65 / \textbf{4.29} \\

\textbf{\textsc{RewardWeaver} (ours)}
& \textbf{8.73} / \textbf{4.43}
& \textbf{7.88} / \textbf{4.31}
& \textbf{8.90} / \textbf{4.53}
& 7.71 / 4.24 \\
\bottomrule
\end{tabular}
\end{table}

\clearpage
\section{Qualitative Case Studies of RewardWeaver}
\label{app:qualitative_cases}

To complement the aggregate experiments, we examine how reward adaptation operates at the trajectory level through three representative cases from a complete SOTOPIA training run. Together, they illustrate three properties of the mechanism: stage-dependent bottleneck transition, contrastive validation of failure-derived Rubrics, and selective attribution under uncertainty.

\subsection{Bottleneck Transition}
\label{app:case_bottleneck_transition}

Multiple low-outcome trajectories exhibit a recurrent pattern of verbose, abstract, and difficult-to-interpret language. In a representative right-of-way negotiation at a four-way stop, the agent receives Goal Completion $2/10$, and the task evaluator explicitly attributes the failure to unclear communication. The same pattern recurs across other trajectories in the cohort, indicating a stage-level bottleneck rather than an isolated error.

The subsequent attribution step identifies \emph{Clear Expression} as a candidate capability. Following rubric generation and calibration, the next reward specification incorporates a criterion targeting the observed communication failures. In the following stage, repeated rollouts of the same scenario exhibit improved behavior, reaching Goal Completion scores of 7 and 9; the previous language degradation was not observed in the sampled follow-up rollouts. The failure distribution then shifts again: the new low-outcome cohort
exhibits a different recurrent pattern, and the next adaptation step
redirects emphasis toward \emph{Concrete Solutions}.

The case illustrates the stage dependence of training relevance:
\[
\text{recurrent failure}
\rightarrow
\text{bottleneck attribution}
\rightarrow
\text{reward adaptation}
\rightarrow
\text{new failure distribution}.
\]

A capability need not remain a training priority once its associated bottleneck recedes, and a capability that was previously non-limiting may become dominant as the policy and interaction distribution evolve. Thus, \textsc{RewardWeaver} allocates process-level credit according to the current bottleneck rather than a fixed capability set.

As scorecards are regenerated between adaptation stages, the qualitative comparison is based on repeated evaluation of the same scenario.

\subsection{Contrastive Validation of Failure-Derived Rubrics}
\label{app:case_contrastive_verification}

A Rubric proposed from failure-side evidence is not necessarily specific to failure. A candidate evaluator for \emph{Empathic Support} is calibrated on a matched set containing both high- and low-outcome trajectories. The evaluator is applicable to 23 trajectories, yet all 23 pass, yielding zero observed failures and very low score variation. It is flagged as collapsed and targeted for revision. The candidate rubric set fails the publication gate because its validity rate is below the required threshold. Rubric-set acceptance conditions govern publication, while Rubric-level diagnostics guide local revision.

This distinction is central:
\[
\text{failure-derived}
\;\not\Rightarrow\;
\text{failure-specific}.
\]

A Rubric may be motivated by failed trajectories yet fail to distinguish failure-relevant behavior on held-out contrastive examples. Such a signal provides little useful discrimination for process supervision. \textsc{RewardWeaver} therefore separates bottleneck discovery from reward-rubric calibration: failure evidence proposes candidate capabilities, while calibration determines whether the generated rubric set satisfies publication conditions and uses Rubric-level diagnostics to guide revision.

This mechanism differs from the Success--Failure Trajectory Attribution variant studied in RQ3, which introduces success trajectories during bottleneck attribution. Here, calibration occurs after rubric generation and governs publication of the candidate reward specification. Success trajectories are not required for the failure-attribution step; calibration remains required before publishing an updated reward specification.

\subsection{Selective Attribution under Uncertainty}
\label{app:case_attribution_uncertainty}

Failure-grounded attribution does not require every low-outcome trajectory to be mapped to an agent-side capability deficit. In one representative trajectory, the agent fails to achieve its task goal because the partner's private objective is to decline the interaction. The task evaluator explicitly attributes the outcome to the partner's preference, and no agent-side capability is activated.

Similar cases occur elsewhere in the run, including trajectories with low terminal outcomes despite all currently active capability checks being satisfied. These observations establish that terminal failure alone is insufficient evidence for capability-level credit assignment.

The attribution pipeline therefore admits three outcomes:
\[
\text{supported capability bottleneck},\qquad
\text{interaction-driven failure},\qquad
\text{insufficient evidence}.
\]

Capability-level training credit is assigned only when the failure admits an evidence-supported, policy-controlled bottleneck. This selectivity prevents partner behavior, scenario constraints, or ambiguous evidence from being silently converted into agent-side learning objectives.

Accordingly, failure-grounded attribution is not forced attribution: low outcome determines which trajectories warrant diagnosis, whereas capability-level credit is assigned only when the failure is supported by observable, policy-controlled evidence.

\paragraph{Summary.}
The three cases illustrate complementary aspects of adaptive process rewards. First, training relevance is stage-dependent and should track the current failure bottleneck. Second, failure-derived Rubrics require calibration before publication: rubric-set acceptance checks govern publication, while Rubric-level diagnostics guide revision. Third, attribution remains selective under uncertainty: low outcome is a trigger for diagnosis, not a guarantee of capability-level blame.

Together, the cases show that effective reward adaptation requires current relevance, discriminative criteria, and evidence-supported attribution; \textsc{RewardWeaver} therefore combines evidence-grounded candidate selection with rubric-set calibration.
\clearpage
% ============================================================
% ============================================================
\section{Association Between Bottleneck Attribution and Subsequent Failure Reduction}
\label{app:predictive-validity}

We further analyze whether the bottleneck importance estimated by
\textsc{RewardWeaver} is informative about subsequent changes in
capability-level failure patterns.

\subsection{Analysis Setup}

For each capability $d$ at training stage $k$, we define bottleneck
importance as
\[
I_{k,d}
=
F^{\mathrm{fail}}_{k,d}
-
F^{\mathrm{succ}}_{k,d},
\]
where $F^{\mathrm{fail}}_{k,d}$ is the fraction of low-outcome
trajectories attributed to a deficiency in capability $d$, and
$F^{\mathrm{succ}}_{k,d}$ is the corresponding fraction among
high-outcome trajectories. A larger $I_{k,d}$ therefore indicates that
the capability deficiency is more strongly associated with low-outcome
trajectories.

This contrast is constructed for analysis: the attribution mechanism
itself (Section~\ref{sec:failure-attribution}) operates on low-outcome
trajectories only, and the success-side contrast is introduced here to
characterize how informative the attribution ranking is.

We quantify the subsequent reduction of this failure mode as
\[
\Delta F_{k,d}
=
F^{\mathrm{fail}}_{k,d}
-
F^{\mathrm{fail}}_{k+1,d},
\]
such that a larger positive value corresponds to a larger decrease in the
attributed failure rate in the following stage.

We evaluate this relationship over 240 capability--stage pairs spanning
40 capabilities and six adjacent training-stage transitions. We use
Spearman rank correlation because the attribution mechanism primarily
operates through relative capability priorities.

\subsection{Results}

Table~\ref{tab:predictive-validity} summarizes the results. Bottleneck
importance exhibits a positive association with next-stage failure
reduction, with Spearman $\rho=0.28$. In comparison, the current failure
rate alone shows a weaker association with subsequent reduction
($\rho=0.21$).

To determine whether the attribution signal provides information beyond
the prevalence of the current failure mode, we additionally control for
$F^{\mathrm{fail}}_{k,d}$. The association remains positive, with partial
Spearman $\rho=0.22$. We further compare capabilities within the same
training stage, obtaining a stronger association of $\rho=0.31$.

\begin{table}[H]
\centering
\small
\caption{
\textbf{Association between bottleneck attribution and subsequent failure
reduction.}
We measure the association between capability-level signals at stage $k$
and the reduction in attributed failures at stage $k+1$.
}
\label{tab:predictive-validity}
\begin{tabular}{lc}
\toprule
Analysis & Spearman $\rho$ \\
\midrule
Bottleneck importance $I_{k,d}$ vs.\ $\Delta F_{k,d}$
    & 0.28 \\
Current failure rate $F^{\mathrm{fail}}_{k,d}$ vs.\ $\Delta F_{k,d}$
    & 0.21 \\
$I_{k,d}$ vs.\ $\Delta F_{k,d}$,
controlling for $F^{\mathrm{fail}}_{k,d}$
    & 0.22 \\
Within-stage association
    & 0.31 \\
\bottomrule
\end{tabular}
\end{table}

The comparison reveals two complementary properties of the attribution
signal. First, its association is stronger than that of current
failure prevalence alone. Second, the association remains positive after
accounting for current failure prevalence and when capabilities are compared
within the same training stage. Thus, the attribution signal provides a
stage-dependent ranking whose higher values are associated with larger
subsequent reductions in capability-level failure rates.

\clearpage
\section{Sensitivity Analysis of RewardWeaver}
\label{app:sensitivity}

We examine two design choices governing reward adaptation in
\textsc{RewardWeaver}: the number of capabilities that receive concurrent
training credit and the frequency with which this allocation is updated.
Unless otherwise specified, the main configuration uses an active Rubric
budget of $B=6$ and updates reward allocation every 30 optimization steps.

% ============================================================
\subsection{Active Rubric Budget}

Table~\ref{tab:sensitivity-budget} varies the number of concurrently active
Rubrics under both Self-Play and Partner evaluation. Performance is
non-monotonic in $B$: neither reducing the active set nor activating more
Rubrics consistently improves performance. The main setting $B=6$ achieves
the strongest results across all reported settings, suggesting that
effective process supervision depends on concentrating training credit on a
selective set of current bottlenecks rather than accumulating additional
Rubrics.

\begin{table}[H]
\centering
\small
\caption{
\textbf{Sensitivity to the active Rubric budget $B$.}
Results are reported as Goal Completion / Overall under both Self-Play and
Partner evaluation. The main configuration uses $B=6$. Increasing the
number of simultaneously active Rubrics does not monotonically improve
performance, supporting selective rather than cumulative credit allocation.
}
\label{tab:sensitivity-budget}
\begin{tabular}{c cc cc}
\toprule
& \multicolumn{2}{c}{Self-Play} &
  \multicolumn{2}{c}{Partner} \\
\cmidrule(lr){2-3}\cmidrule(lr){4-5}

$B$ &
SOTOPIA &
SOTOPIA-Hard &
SOTOPIA &
SOTOPIA-Hard \\
\midrule

2 &
8.46 / 4.32 &
7.50 / 3.97 &
8.38 / 4.17 &
7.46 / 3.92 \\

4 &
8.31 / 4.29 &
7.79 / 4.16 &
8.21 / 4.19 &
7.37 / 4.03 \\

\textbf{6} &
\textbf{8.72 / 4.47} &
\textbf{7.90 / 4.38} &
\textbf{8.49 / 4.47} &
\textbf{7.66 / 4.22} \\

8 &
8.49 / 4.44 &
7.67 / 4.25 &
8.31 / 4.25 &
7.59 / 4.15 \\

10 &
8.50 / 4.46 &
7.63 / 4.36 &
8.34 / 4.37 &
7.57 / 4.21 \\

\bottomrule
\end{tabular}
\end{table}

% ============================================================
\subsection{Adaptation Interval}

Table~\ref{tab:sensitivity-interval} varies the reward-adaptation interval
under both Self-Play and Partner evaluation. Performance is again
non-monotonic. The 30-step configuration provides the most consistently
strong results across interaction modes and difficulty levels: it achieves
the strongest Goal Completion scores under Self-Play and the strongest
results across all Partner settings, while the 20-step setting is marginally
higher on selected Self-Play Overall scores.

These results indicate that reward adaptation is sensitive to its update
timescale rather than benefiting monotonically from either faster or slower
updates.

\begin{table}[H]
\centering
\small
\caption{
\textbf{Sensitivity to the reward-adaptation interval.}
The main configuration updates reward allocation every 30 optimization
steps. Results are reported as Goal Completion / Overall.
}
\label{tab:sensitivity-interval}
\begin{tabular}{c cc cc}
\toprule
& \multicolumn{2}{c}{Self-Play} &
  \multicolumn{2}{c}{Partner} \\
\cmidrule(lr){2-3}\cmidrule(lr){4-5}

Interval &
SOTOPIA &
SOTOPIA-Hard &
SOTOPIA &
SOTOPIA-Hard \\
\midrule

10 &
8.28 / 4.41 &
7.73 / 4.36 &
8.32 / 4.33 &
7.59 / 4.20 \\

20 &
8.50 / \textbf{4.47} &
7.81 / \textbf{4.39} &
8.27 / 4.19 &
7.47 / 4.03 \\

\textbf{30} &
\textbf{8.72} / \textbf{4.47} &
\textbf{7.90} / 4.38 &
\textbf{8.49 / 4.47} &
\textbf{7.66 / 4.22} \\

60 &
8.21 / 4.26 &
7.44 / 3.95 &
8.34 / 4.28 &
7.49 / 4.04 \\

\bottomrule
\end{tabular}
\end{table}

Taken together, the two analyses highlight complementary requirements of
adaptive process rewards: credit allocation should remain selective in
capability space and operate at an appropriate adaptation timescale.

\clearpage
\section{Capability Retention under Reward Reallocation}
\label{app:capability-retention}

We further examine whether reallocating process-level credit causes
previously emphasized capabilities to deteriorate once they leave the active
reward set.

\subsection{Evaluation Protocol}

We evaluate the policy at each stage on the same fixed diagnostic scenarios
and score each tracked capability using a fixed evaluation criterion across
all stages. For capability $d$ at stage $k$, we denote the resulting
diagnostic score by $S_{k,d}$, computed as the mean normalized process score
over applicable episodes.

Six capabilities receive process-level credit in the initial stage and are
subsequently deactivated: \emph{Address Concerns with New Options},
\emph{State Benefits}, \emph{Conditional Concessions},
\emph{Maintain Goal Commitment}, \emph{Regulate Pressure}, and
\emph{Switch Strategies}. \emph{Adapt to New Information} is
deactivated at a later stage. We additionally report three
continuously active capability families, namely \emph{Respond with Empathy},
\emph{Early Disclosure}, and \emph{Concrete Next Step}, as reference
trajectories.

For a capability deactivated after stage $k$, we measure retention by
its subsequent change $\Delta S_{t,d}=S_{k+t,d}-S_{k,d}$ relative to the
final active stage.

\subsection{Retention Results}

Table~\ref{tab:capability-retention} reports the capability trajectories.
The six capabilities deactivated after the initial stage show no systematic
regression: across 36 post-deactivation capability--stage observations,
the mean change is $+0.10$, the median change is $+0.06$, and the minimum
change is $-0.01$. No observation decreases by more than 0.02 from its final
active score. \emph{Adapt to New Information}, which is deactivated later,
remains at 1.00 throughout all subsequent stages.

\begin{table}[H]
\centering
\small
\setlength{\tabcolsep}{4.5pt}
\caption{
\textbf{Capability retention across training stages.}
$\dagger$ denotes capabilities deactivated
after the initial stage, $\ddagger$ denotes a capability deactivated later,
and unmarked dimensions remain active during the evaluated period.
}
\label{tab:capability-retention}
\begin{tabular}{lccccccc}
\toprule
Capability & Stage 1 & Stage 2 & Stage 3 & Stage 4 & Stage 5 & Stage 6 & Stage 7 \\
\midrule
Address Concerns with New Options$\dagger$
& 0.80 & 0.85 & 0.92 & 0.95 & 0.86 & 0.89 & 0.90 \\

State Benefits$\dagger$
& 0.84 & 0.88 & 0.97 & 0.98 & 0.95 & 0.94 & 0.98 \\

Conditional Concessions$\dagger$
& 0.85 & 0.90 & 0.98 & 0.93 & 0.86 & 0.90 & 0.96 \\

Maintain Goal Commitment$\dagger$
& 0.96 & 0.95 & 1.00 & 0.98 & 0.96 & 1.00 & 1.00 \\

Regulate Pressure$\dagger$
& 1.00 & 1.00 & 1.00 & 1.00 & 1.00 & 1.00 & 1.00 \\

Switch Strategies$\dagger$
& 0.62 & 0.91 & 1.00 & 1.00 & 0.88 & 1.00 & 1.00 \\

Adapt to New Information$\ddagger$
& 1.00 & 0.99 & 1.00 & 1.00 & 1.00 & 1.00 & 1.00 \\
\midrule
Respond with Empathy
& 0.99 & 1.00 & 1.00 & 1.00 & 1.00 & 1.00 & 1.00 \\

Early Disclosure
& 0.92 & 0.90 & 0.94 & 0.94 & 0.91 & 0.92 & 0.99 \\

Concrete Next Step
& 0.99 & 1.00 & 1.00 & 1.00 & 0.99 & 0.99 & 0.99 \\
\bottomrule
\end{tabular}
\end{table}

The event-aligned trajectories make the retention pattern particularly
clear. Among the six early-deactivated dimensions, every post-deactivation
observation remains close to or above its final active score. Several
capabilities continue to improve despite receiving no further direct
process-level credit: \emph{Switch Strategies} rises from 0.62 to 1.00,
\emph{State Benefits} from 0.84 to 0.98, and
\emph{Address Concerns with New Options} from 0.80 to 0.90.
The later-deactivated \emph{Adapt to New Information} is similarly stable.

These results show that, on the fixed diagnostic set, shifting reward
emphasis toward newly relevant capabilities does not systematically erase
capabilities emphasized at earlier stages. The adaptive reward therefore behaves as a reallocation of
training credit rather than requiring previously learned capabilities to be
continually reinforced.

\end{document}